\documentclass[sigconf]{acmart}

\usepackage{multirow}
\usepackage[utf8]{inputenc} 
\usepackage{url}            
\usepackage{booktabs}       
\usepackage{amsfonts}       
\usepackage{nicefrac}       
\usepackage{xcolor}         
\usepackage{colortbl}       
\usepackage{float}
\usepackage{amsthm}
\usepackage{latexsym}
\usepackage{thmtools}
\usepackage{thm-restate}
\usepackage{etoolbox}
\usepackage{array}
\usepackage{makecell}
\usepackage{tabularx}
\usepackage{enumitem}
\usepackage{titlesec}
\usepackage{subcaption}
\usepackage{comment}
\usepackage{algorithm}
\usepackage[noend]{algpseudocode}
\usepackage{pifont}
\usepackage[framemethod=TikZ]{mdframed}
\usepackage{bbding}
\usepackage{verbatim,ifthen,alltt}
\usepackage{microtype}      
\usepackage{stmaryrd}
\usepackage{xfrac}
\usepackage{xspace}
\usepackage{tikz}
\usepackage{pgf}
\usepackage{forest}

\usetikzlibrary{fit,calc,tikzmark}
\usetikzlibrary{arrows,decorations.pathmorphing,decorations.pathreplacing,decorations.footprints,fadings,calc,trees,mindmap,shadows,decorations.text,patterns,positioning,shapes}
\usetikzlibrary{arrows,shadows,backgrounds}
\usetikzlibrary{arrows.meta}
\usetikzlibrary{positioning,fit,automata,matrix}
\usetikzlibrary{shapes.symbols,shapes.misc,shapes.arrows}

\definecolor{darkred}{rgb}{0.7,0.1,0.1}
\definecolor{medred}{rgb}{0.5,0.1,0.1}
\definecolor{midred}{rgb}{0.7,0.2,0.2}
\definecolor{vdarkred}{rgb}{0.4,0.1,0.1}
\definecolor{darkslategray}{rgb}{0.18, 0.31, 0.31} 
\definecolor{platinum}{rgb}{0.9, 0.89, 0.89} 
\definecolor{gray}{rgb}{.4,.4,.4}
\definecolor{midgrey}{rgb}{0.5,0.5,0.5}
\definecolor{middarkgrey}{rgb}{0.35,0.35,0.35}
\definecolor{darkgrey}{rgb}{0.3,0.3,0.3}
\definecolor{darkred}{rgb}{0.7,0.1,0.1}
\definecolor{midblue}{rgb}{0.2,0.2,0.7}
\definecolor{darkblue}{rgb}{0.1,0.1,0.5}
\definecolor{darkgreen}{rgb}{0.1,0.5,0.1}
\definecolor{defseagreen}{cmyk}{0.69,0,0.50,0}
\definecolor{purple3}{RGB}{125,38,205}          

\definecolor{tyellow1}{HTML}{FCE94F}
\definecolor{tyellow2}{HTML}{EDD400}
\definecolor{tyellow3}{HTML}{C4A000}
\definecolor{torange1}{HTML}{FCAF3E}
\definecolor{torange2}{HTML}{F57900}
\definecolor{torange3}{HTML}{C35C00}
\definecolor{tbrown1}{HTML}{E9B96E}
\definecolor{tbrown2}{HTML}{C17D11}
\definecolor{tbrown3}{HTML}{8F5902}
\definecolor{tgreen1}{HTML}{8AE234}
\definecolor{tgreen2}{HTML}{73D216}
\definecolor{tgreen3}{HTML}{4E9A06}
\definecolor{tblue1}{HTML}{729FCF}
\definecolor{tblue2}{HTML}{3465A4}
\definecolor{tblue3}{HTML}{204A87}
\definecolor{tpurple1}{HTML}{AD7FA8}
\definecolor{tpurple2}{HTML}{75507B}
\definecolor{tpurple3}{HTML}{5C3566}
\definecolor{tred1}{HTML}{EF2929}
\definecolor{tred2}{HTML}{CC0000}
\definecolor{tred3}{HTML}{A40000}
\definecolor{tlgray1}{HTML}{EEEEEC}
\definecolor{tlgray2}{HTML}{D3D7CF}
\definecolor{tlgray3}{HTML}{BABDB6}
\definecolor{tdgray1}{HTML}{888A85}
\definecolor{tdgray2}{HTML}{555753}
\definecolor{tdgray3}{HTML}{2E3436}

\newcommand{\dghlight}[1]{{\color[RGB]{0,120,0}#1}}

\usepackage[noabbrev,nameinlink,capitalise]{cleveref}

\AtBeginDocument{%
  }

\setcopyright{acmcopyright}
\copyrightyear{2025}
\acmYear{2025}
\acmDOI{XXXXXXX.XXXXXXX}

\acmConference[arXiv]{arXiv Report}{January, 2025}{TbD}

\makeatletter
\def\thm@space@setup{\thm@preskip=5.0pt
\thm@postskip=0pt}
\makeatother

\newtheoremstyle{newstyle}      
{2.5pt} 
{2.5pt} 
{\mdseries} 
{} 
{\bfseries} 
{.} 
{ } 
{} 
\theoremstyle{newstyle}

\newtheorem{proposition}{Proposition}

\newtheorem{example}{Example}

\crefname{theorem}{Theorem}{Theorems}
\crefname{lemma}{Lemma}{Lemmas}
\crefname{proposition}{Proposition}{Propositions}
\crefname{definition}{Definition}{Definitions}
\crefname{corollary}{Corollary}{Corollaries}
\crefname{example}{Example}{Examples}
\crefname{claim}{Claim}{Claims}
\crefname{assumption}{Assumption}{Assumptions}
\crefname{enumi}{}{}



\newcommand{\cmark}{\mbox{\ding{51}}}
\newcommand{\xmark}{\mbox{\ding{55}}}

\newcommand{\nfrac}{\nicefrac}

\newcommand{\fml}[1]{{\mathcal{#1}}}

\newcommand{\tn}[1]{\textnormal{#1}}

\newcommand{\mbf}[1]{\ensuremath\mathbf{#1}}
\newcommand{\msf}[1]{\ensuremath\mathsf{#1}}
\newcommand{\mbb}[1]{\ensuremath\mathbb{#1}}

\newcommand{\waxp}{\ensuremath\mathsf{WAXp}}
\newcommand{\wcxp}{\ensuremath\mathsf{WCXp}}

\newcommand{\sbwaxp}{\ensuremath\mathsf{sbWAXp}}

\newcommand{\cgtalg}{\ensuremath\mathsf{CGT}}

\newcommand{\shap}[1]{\ensuremath\msf{SHAP}_{\mathrm{#1}}}
\newcommand{\nshap}[1]{\ensuremath\nu\shap{#1}}
\newcommand{\sv}{\ensuremath\msf{Sc}}
\newcommand{\ssc}[1]{\ensuremath\mathrm{SHAP}^{#1}} 

\newcommand{\similar}{\ensuremath\sigma}
\newcommand{\tsimilar}{\ensuremath\mathsf{T}\sigma}
\newcommand{\simrelop}{\ensuremath\approx}

\newcommand{\critic}{\ensuremath\msf{Crit}}

\newcounter{tableeqn}[table]

\DeclareMathOperator*{\limply}{\rightarrow}

\newcommand{\exv}{\ensuremath\mathbf{E}}

\newcommand{\prob}{\ensuremath\mathbf{Pr}}

\newcommand{\cf}{\ensuremath\upsilon} 
\newcommand{\cfn}[1]{\ensuremath\upsilon_{#1}} 

\newcommand{\svn}[1]{\msf{Sc}_{#1}}

\newcommand{\shv}[1]{\ensuremath\msf{Sc}_{#1}}
\newcommand{\she}[1]{\ensuremath\msf{\hat{Sc}}_{#1}}
\newcommand{\shany}{\ensuremath\msf{Sc}^{\ast}}
\newcommand{\rnk}[1]{\ensuremath\msf{Rank_{#1}}}

\newcommand{\jnote}[1]{\medskip\noindent$\llbracket$\textcolor{darkred}{joao}: \emph{\textcolor{middarkgrey}{#1}}$\rrbracket$\medskip}

\newcommand{\jnoteF}[1]{}

\newcolumntype{L}[1]{>{\raggedright\let\newline\\\arraybackslash\hspace{0pt}}m{#1}}
\newcolumntype{C}[1]{>{\centering\let\newline\\\arraybackslash\hspace{0pt}}m{#1}}
\newcolumntype{R}[1]{>{\raggedleft\let\newline\\\arraybackslash\hspace{0pt}}m{#1}}

\tikzset{
  0 my edge/.style={densely dashed, my edge, draw=midblue},
  my edge/.style={-{Stealth[]}, draw=midblue},
}

\def\HiLi{\leavevmode\rlap{\hbox to \linewidth{\color{platinum}\leaders\hrule height .8\baselineskip depth .5ex\hfill}}}

\newlength{\Oldarrayrulewidth}

\titlespacing{\section}{0pt}{*2.0}{*1.0}
\titlespacing{\subsection}{0pt}{*1.25}{*0.75}
\titlespacing{\subsubsection}{0pt}{*0.35}{*0.5}
\titlespacing{\paragraph}{0pt}{*0.275}{*0.575}
\titleformat{\paragraph}[runin]
  {\bfseries}{\theparagraph}{1em}{}
            
\makeatletter
\newcommand\nparagraph{%
  \@startsection{paragraph}
    {4}
    {\z@}
    {0.225ex \@plus0.225ex \@minus.125ex}
    {-1em}
    {\normalfont\normalsize\bfseries}%
}
\makeatother

\setitemize{noitemsep,topsep=0pt,parsep=0pt,partopsep=0pt}
\setenumerate{noitemsep,topsep=0pt,parsep=0pt,partopsep=0pt}
\setlist{nosep,leftmargin=0.45cm}

\AtBeginDocument{%
  \setlength{\belowdisplayskip}{2.575pt} \setlength{\belowdisplayshortskip}{2.75pt}
  \setlength{\abovedisplayskip}{2.575pt} \setlength{\abovedisplayshortskip}{2.75pt}
}

\providetoggle{long}
\settoggle{long}{false}

\makeatletter
\algnewcommand{\LineComment}[1]{\Statex \hskip\ALG@thistlm \(\triangleright\) #1}
\makeatother

\newcommand{\iter}{\ensuremath\mathsf{iter}}
\newcommand{\perm}{\ensuremath\fml{P}}
\newcommand{\pref}{\ensuremath\msf{Pref}}
\newcommand{\True}{\textbf{true}}
\newcommand{\False}{\textbf{false}}
\newcommand{\kwnot}{\textbf{not}\xspace}
\newcommand{\kwand}{\textbf{and}\xspace}
\newcommand{\kwdownto}{\textbf{downto}\xspace}

\begin{document}

\title[The Explanation Game -- Rekindled]{The Explanation Game -- Rekindled\\(Extended Version)}

\author{Joao Marques-Silva}
\authornote{All authors contributed equally to this research.}
\email{jpms@icrea.cat}
\orcid{0000-0002-6632-3086}
\affiliation{%
  \institution{ICREA, Univ.\ Lleida}
  \city{Lleida}
  \country{Spain}
  \postcode{25001 Lleida}
}

\author{Xuanxiang Huang}
\authornotemark[1]
\email{xuanxiang.huang@cnrsatcreate.sg}
\orcid{0000-0002-3722-7191}
\affiliation{%
  \institution{CNRS@CREATE}
  \city{Singapore}
  \country{Singapore}
  \postcode{31400}
}

\author{Olivier L\'{e}toff\'{e}}
\authornotemark[1]
\email{olivier.letoffe@orange.fr}
\affiliation{%
  \institution{IRIT, Univ.\ Toulouse}
  \city{Toulouse}
  \country{France}
  \postcode{31400}
}








\renewcommand{\shortauthors}{Marques-Silva, Huang and L\'{e}toff\'{e}}

\begin{abstract}
  Recent work demonstrated the existence of critical flaws in the
  current use of Shapley values in explainable AI (XAI), i.e.\ the
  so-called SHAP scores.
  These flaws are significant in that the scores provided to a
  human decision-maker can be misleading.
  Although these negative results might appear to indicate that
  Shapley values ought not be used in XAI,
  this paper argues otherwise.
  %
  Concretely, this paper 
  proposes a novel definition of SHAP scores that overcomes existing
  flaws.
  Furthermore, the paper outlines a practically efficient solution for
  the rigorous estimation of the novel SHAP scores.
  Preliminary experimental results confirm our claims, and further
  underscore the flaws of the current SHAP scores. 
  %
\end{abstract}

\ccsdesc[500]{Computing methodologies~Artificial intelligence}
\ccsdesc[500]{Computing methodologies~Machine learning algorithms}
\ccsdesc[500]{Theory of computation~Automated reasoning}
\ccsdesc[500]{Computing methodologies~Machine learning}

\keywords{Explainable AI, Shapley values, Abductive reasoning}


\maketitle

\section{Introduction} \label{sec:intro}

The importance of explainable artificial intelligence (XAI) cannot be
overstated, being widely accepted as a fundamental pillar of
trustworthy AI.
Both in the European Union (EU) and the United States (US), there has
been recent legislation aiming at regulating the use of systems of AI,
especially in domains that directly impact humans.%
\footnote{\url{https://www.nist.gov/artificial-intelligence/executive-order-safe-secure-and-trustworthy-artificial-intelligence},
Oct.~2023.}%
\textsuperscript{,}%
\footnote{\url{https://artificialintelligenceact.eu/}, Jun.~2024.}
%
Nevertheless, some experts have warned of existential risks caused by
AI.%
\footnote{\url{https://www.bbc.com/news/uk-65746524}, May.~2023.}
Similarly, recent studies contemplate existential risks caused by
the advances in AI.%
\footnote{\url{https://www.gladstone.ai/action-plan}, Feb.~2024.}
%
%
As a result,
XAI is expected to provide critical support in attaining 
much-needed trust in systems of AI.

XAI methods can be broadly categorized into those based on feature
attribution and those based on feature selection. Feature attribution
aims at assigning degrees of importance to features, thereby enabling
the ranking of the relative importance of features. In contrast,
feature selection aims at selecting subsets of features, each encoding
a rule for some prediction. For example, Anchors~\cite{guestrin-aaai18}
exemplifies XAI by feature selection, while LIME~\cite{guestrin-kdd16}
exemplifies XAI by feature attribution. Moreover, whereas some works
are model-agnostic~\cite{guestrin-kdd16,guestrin-aaai18}, others
require access to the machine learning (ML)
models~\cite{muller-plosone15}. Despite the importance of XAI, most
past efforts are characterized by lack of 
rigor~\cite{ignatiev-ijcai20,ms-isola24,hermanns-acml24}.
As a consequence, most methods of XAI in current use are unworthy of
trust~\cite{wjgms-aimag24}.

%
The tool SHAP~\cite{lundberg-nips17} is arguably the most popular
method of XAI by feature attribution. 
As its core, the tool SHAP approximates Shapley
values~\cite{shapley-ctg53} based on a definition that is standard in
XAI~\cite{kononenko-jmlr10,kononenko-kis14}.
Shapley values are defined on a game, which consists of a set of
elements (e.g.\ voters) and a given characteristic function that maps
subsets of a set of elements (e.g.\ players or voters in game theory,
or features in an ML model) to the reals.
Depending on the characteristic function used, different Shapley
values are obtained. However, for all chosen characteristic functions,
the obtained values respect the key axioms of Shapley
values~\cite{shapley-ctg53}.
In the case of game theory, concretely when measuring a priori voting
power, Shapley values are instantiated starting from a given
characteristic function, which was first proposed in 1954 by Shapley
\& Shubik~\cite{shapley-apsr54}.
The same characteristic function has been used, explicitly or
implicitly, in most other proposals of power indices, i.e. measures of
relative voting power. This is the case with the well-known Banzhaf,
Deegan-Packel and Holler-Packel
indices~\cite{machover-bk98,lhams-corr24}, among many others. 

In the case of XAI, the tool SHAP is based on a concrete instantiation
of Shapley values, one that uses a specific characteristic function.
However, this characteristic function bears no relationship with those
that have been used in a priori voting power since the
1950s~\cite{machover-bk98}.
The theoretical SHAP scores denote the values that the tool 
SHAP only approximates, given the chosen characteristic function.
%
%
Since its publication in 2017, the tool SHAP, and the corresponding
estimated SHAP scores, have become ubiquitous in a growing number of
uses of XAI.
Furthermore, the complexity of computing the theoretical SHAP scores 
has been studied in a number of
works~\cite{vandenbroeck-jair22,barcelo-jmlr23}. 

Our recent work~\cite{msh-cacm24} demonstrated
that theoretical SHAP scores can produce misleading
information,\footnote{%
The theoretical SHAP scores are the values that the tool SHAP
would produce if it were capable of computing SHAP scores with the
utmost rigor.
Our results apply to the theory of SHAP scores, and as a consequence
to the values that the tool SHAP approximates. The corollary is that
the tool SHAP approximates Shapley values that can be misleading for a
human-decision maker.}
%
assigning the most importance to absolutely unimportant features,
and not assigning any importance to critically important features.
These negative results 
are further corroborated by several additional cases
studies~\cite{hms-corr23a,hms-ijar24}.
It should be plain that 
the results in~\cite{msh-cacm24} also serve to 
question the conclusions from
thousands of papers that build on SHAP, 
but also from the massive uses of SHAP in companies 
and 
in academic and research institutions.
Motivated by the flaws in SHAP's theoretical foundations, some
researchers have proposed explanations by feature attribution 
based on exploiting alternatives to Shapley
values~\cite{izza-aaai24,ignatiev-sat24}. The downside of such 
approaches is that these are \emph{not} based on Shapley values, and
compute measures of importance that can also yield unsatisfactory
results~\cite{lhams-corr24}.
Furthermore, as pointed out in recent
work~\cite{lhms-aaai25,lhams-corr24}, 
the limitations of SHAP scores are not the result of Shapley values
per se, but are instead the result of the characteristic function that
has become ubiquitous in XAI. Finally, as also argued%
in
~\cite{lhms-aaai25,lhams-corr24}, 
other characteristic functions in use, 
e.g.\ baselines~\cite{najmi-icml20}, are also inadequate.

Given the above, a natural question is whether the use of SHAP scores
should be relinquished, at least in application domains where the rigor
of explanations is paramount. Although the answer must be affirmative
for existing implementations (and supporting theory) of SHAP
scores~\cite{lundberg-nips17}, this paper proposes a rigorous solution
for the computation of Shapley values for XAI (we will continue to use
the name \emph{SHAP scores}) such that all the limitations reported in
earlier work~\cite{msh-cacm24} are overcome. 
Moreover, the proposed solution further highlights the connections
between explanations using feature attribution and explanations using
feature selection, which are revealed by adopting a logic-based
approach to explainability.
Finally, the experiments provide conclusive practical evidence to the
flaws of both of earlier definitions of SHAP scores and the tool SHAP.

\section{Definitions} \label{sec:prelim}

This section adapts and extends the most recent notation used in the
field of logic-based explainability~\cite{ms-isola24}.

\paragraph{Classification \& regression problems.}
%
Let $\fml{F}=\{1,\ldots,m\}$ denote a set of features.
Each feature $i\in\fml{F}$ takes values from a domain $\mbb{D}_i$.
Domains can be categorical or ordinal. If ordinal, domains can be
discrete or real-valued. 
Feature space is defined by
$\mbb{F}=\mbb{D}_1\times\mbb{D}_2\times\ldots\times\mbb{D}_m$. 
Throughout the paper domains are assumed to be discrete-valued.
%
Thus, for real-valued features, some sort of finite discretization is
assumed.%
\footnote{Discretization of real-valued domains is fairly
standard, e.g.~\cite{guestrin-aaai18}.}
%
%
The notation $\mbf{x}=(x_1,\ldots,x_m)$ denotes an arbitrary point in 
feature space, where each $x_i$ is a variable taking values from
$\mbb{D}_i$. Moreover, the notation $\mbf{v}=(v_1,\ldots,v_m)$
represents a specific point in feature space, where each $v_i$ is a
constant representing one concrete value from $\mbb{D}_i$.
A classifier maps each point in feature space to a class taken from
$\fml{K}=\{c_1,c_2,\ldots,c_K\}$. Classes can also be categorical or
ordinal. However, and unless otherwise stated, classes are assumed to
be ordinal.
In the case of regression, each point in feature space is mapped to an
ordinal value taken from a set of values $\mbb{V}$, e.g.\ $\mbb{V}$
could denote $\mbb{Z}$ or $\mbb{R}$.
Therefore, a classifier $\fml{M}_{C}$ is characterized by a
non-constant \emph{classification function} $\kappa$ that maps feature
space $\mbb{F}$ into the set of classes $\fml{K}$,
i.e.\ $\kappa:\mbb{F}\to\fml{K}$.
A regression model $\fml{M}_R$ is characterized by a non-constant
\emph{regression function} $\rho$ that maps feature space $\mbb{F}$
into the set elements from $\mbb{V}$, i.e.\ $\rho:\mbb{F}\to\mbb{V}$. 
A classifier model $\fml{M}_{C}$ is represented by a tuple
$(\fml{F},\mbb{F},\fml{K},\kappa)$, whereas a regression model
$\fml{M}_{R}$ is represented by a tuple
$(\fml{F},\mbb{F},\mbb{V},\rho)$.
When viable, we will represent an ML model $\fml{M}$ by a tuple
$(\fml{F},\mbb{F},\mbb{T},\pi)$, with prediction $\pi:\mbb{F}\to\mbb{T}$,
without specifying whether $\fml{M}$ denotes a classification
or a regression model, and where $\mbb{T}$ can either be $\fml{K}$ or
$\mbb{V}$. 
An instance denotes a pair $(\mbf{v},q)$, where
$\mbf{v}\in\mbb{F}$,
$q\in\mbb{T}$, with $q=\pi(\mbf{v})$.
An explanation problem is a tuple $\fml{E}=(\fml{M},(\mbf{v},q))$,
where $\fml{M}$ is some ML model
and $(\mbf{v},q)$ is a target instance. 
%
%

\paragraph{Similarity relational operators.}
%
%
For each feature $i\in\fml{F}$, we will assume a user-specified
similarity relational operator (a predicate)
$\simrelop\,\,:\mbb{D}_i\times\mbb{D}_i\to\{\top,\bot\}$, such that
$x_i\simrelop{v_i}$ holds true ($\top$) if the values $x_i$ and $v_i$
are deemed sufficiently close to each other.
For example, for categorical features we might require that similarity
should mean that $x_i$ and $v_i$ represent the same value. In contrast,
for real-valued features, we might accept a small difference in value,
with different measures of absolute or relative change envisaged.%
\footnote{\url{https://en.wikipedia.org/wiki/Relative_change}.}
Moreover, the similarity relational operator can be generalized as
follows.
Given $\mbf{x},\mbf{v}\in\mbb{F}$, we define similarity between
the features in some set $\fml{S}\subseteq\fml{F}$:
\[
\mbf{x}_{\fml{S}}\simrelop\mbf{v}_{\fml{S}} \quad {\coloneq} \quad
\bigwedge\nolimits_{i\in\fml{S}}x_i\simrelop{v_i}
\]
when $\fml{S}=\fml{F}$, we write $\mbf{x}\simrelop\mbf{v}$, for
simplicity.
%
%
%
%
Finally, a similarity relational operator is also assumed for the
prediction, such that $\pi(\mbf{x})\simrelop\pi(\mbf{v})$ holds when
$\pi(\mbf{x})$ is sufficiently close to $\pi(\mbf{v})$.

\paragraph{Logic-based explainability.}
%
Abductive and contrastive explanations (resp.~AXps/CXps) are examples
of formal explanations for classification
problems~\cite{ms-rw22,darwiche-lics23}. As argued
in recent work~\cite{ms-isola24}, the same concepts can be generalized
to the case of regression problems. The presentation below just
requires a well-defined similarity operator.

A weak (i.e.\ non-subset minimal) abductive explanation (WAXp) denotes
a set of features $\fml{S}\subseteq\fml{F}$, such that for every point
in feature space the ML model output is similar to the given instance:
$(\mbf{v},q)$. 
The condition for a set of features $\fml{S}\subseteq\fml{F}$ to
represent a WAXp (which also defines a corresponding predicate
$\waxp$) is as follows:%
\footnote{%
For simplicity, predicates parameterized by the explanation problem
$\fml{E}=(\fml{M},(\mbf{v},q))$ are also assumed to be parameterized
by both $\fml{M}$, and the definition of $\fml{M}$, and by
$(\mbf{v},q)$.}
\begin{equation} \label{eq:waxp}
\waxp(\fml{S};\fml{E}) ~~ \coloneq ~~
\forall(\mbf{x}\in\mbb{F}).(\mbf{x}_{\fml{S}}\simrelop\mbf{v}_{\fml{S}})\limply(\pi(\mbf{x})\simrelop\pi(\mbf{v}))
\end{equation}
%
%
Furthermore, an AXp is a subset-minimal WAXp.

A weak contrastive explanation (WCXp) denotes a set of features
$\fml{S}\subseteq\fml{F}$, such that there exists some point in
feature space, where only the features in $\fml{S}$ are allowed to
change, that makes the ML model output distinguishable from the given
instance $(\mbf{v},q)$.
The condition for a set of features $\fml{S}\subseteq\fml{F}$ to
represent a WCXp (which also defines a corresponding predicate
$\wcxp$) is as follows:
\begin{equation} \label{eq:wcxp}
\wcxp(\fml{S};\fml{E}) ~~ \coloneq ~~
\exists(\mbf{x}\in\mbb{F}).(\mbf{x}_{\fml{F}\setminus\fml{S}}\simrelop\mbf{v}_{\fml{F}\setminus\fml{S}})\land(\pi(\mbf{x})\not\simrelop\pi(\mbf{v}))
\end{equation}
%
%
Furthermore, a CXp is a subset-minimal WCXp.
Observe that for ML models that compute some function, (W)CXps
can be viewed as implementing actionable
recourse~\cite{spangher-fat19}. Furthermore, (W)CXps formalize the
notion of contrastive explanation discussed
elsewhere~\cite{miller-aij19}.
Finally, feature (ir)relevancy is a key concept, that served to
demonstrate that SHAP scores can yield misleading
information. 
A feature is \emph{relevant} if it is included in at least one AXp;
otherwise it is \emph{irrelevant}. It is well-known that a relevant
feature will also be included in at least one CXp.

\paragraph{Distributions, expected value.}
Throughout the paper, it is assumed a \emph{uniform probability
distribution} on each feature, and such that all features are
independent.
%
Thus, the probability of an arbitrary point in feature space
becomes:
%
\begin{equation}\label{eq:probpt}
  \prob(\mbf{x}) \coloneq \frac{1}{\Pi_{i\in\fml{F}}|\mbb{D}_i|}
\end{equation}
That is, every point in the feature space has the same
probability. (Observe that, since real-valued features are
discretized and domains are finite, then~\eqref{eq:probpt} is
well-defined.)
The \emph{expected value} of an ML model $\pi:\mbb{F}\to\mbb{T}$
is denoted by $\mbf{E}[\pi(\mbf{x})]$. 
%
Furthermore, let
$\exv[\pi(\mbf{x})\,|\,\mbf{x}_{\fml{S}}\simrelop\mbf{v}_{\fml{S}}]$
represent the expected value of $\pi$ over points in feature space
consistent (given $\simrelop$) with the coordinates of $\mbf{v}$
dictated by $\fml{S}$, which is defined as follows:
%
%
\begin{equation} \label{eq:evdef}
  \exv[\pi(\mbf{x})\,|\,\mbf{x}_{\fml{S}}\simrelop\mbf{v}_{\fml{S}}]
  \coloneq\frac{1}{|\{\mbf{x}\,|\,\mbf{x}_{\fml{S}}\simrelop\mbf{v}_{\fml{S}}\}|}
  \sum\nolimits_{\mbf{x},\,\mbf{x}_{\fml{S}}\simrelop\mbf{v}_{\fml{S}}}\pi(\mbf{x})
\end{equation}
%
%

\jnoteF{
  Given $\mbf{z}\in\mbb{F}$ and$\fml{S}\subseteq\fml{F}$, let
  $\mbf{z}_{\fml{S}}$ represent the vector composed of the coordinates
  of $\mbf{z}$ dictated by $\fml{S}$.
  \[
  \exv{\kappa\,|\,\mbf{x}_{\fml{S}}\simrelop\mbf{v}_{\fml{S}}}
  \coloneq\frac{1}{|\Upsilon(\fml{S};\mbf{v})|}
  \sum\nolimits_{\mbf{x}\in\Upsilon(\fml{S};\mbf{v})}\kappa(\mbf{x})
  \]
}

\paragraph{Shapley values \& SHAP scores.}
%
Shapley values were proposed in the context of game theory in the
early 1950s by L.\ S.\ Shapley~\cite{shapley-ctg53}. Shapley values
are defined given some game $G=(N,\cf)$, with $N=\{1,\ldots,n\}$, and 
$\cf$ a \emph{characteristic function}, i.e.\ a real-valued function
defined on the subsets of $N$, $\cf:2^{N}\to\mbb{R}$.%
\footnote{%
$N$ denotes the elements of the game; in our case these are the
features.}
%
%
It is well-known that Shapley values represent the \emph{unique}
function that respects a number of important
axioms~\cite{shapley-ctg53}.

In the context of explainability, Shapley values are most often
referred to as SHAP scores%
~\cite{kononenko-jmlr10,kononenko-kis14,lundberg-nips17},%
\footnote{%
The complexity of computing SHAP scores has been the subject of
several recent
works~\cite{vandenbroeck-jair22,barcelo-jmlr23}.} 
and consider a specific characteristic function
$\cf_e:2^{\fml{F}}\to\mbb{R}$,
which is defined by,
\begin{equation} \label{eq:cfs}
  \cf_e(\fml{S};\fml{E}) \coloneq
  \exv[\pi(\mbf{x})\,|\,\mbf{x}_{\fml{S}}\simrelop\mbf{v}_{\fml{S}}]
\end{equation}
%
%
Thus, given a set $\fml{S}$ of features,
$\cf_e(\fml{S};\fml{E})$ represents the \emph{e}xpected value
of the classifier over the points of feature space
having the features dictated by $\fml{S}$ similar to those of
$\mbf{v}$.
%
%
%
The formulation presented in earlier
work~\cite{barcelo-jmlr23} allows for different input 
distributions when computing the average values. For the purposes of
this paper, it suffices to consider solely a uniform input 
distribution, and so the dependency on the input distribution is not
accounted for.
Independently of the distribution considered, it should be clear that
in most cases $\cfn{e}(\emptyset)\not=0$; this is the case for example
with boolean classifiers~\cite{barcelo-jmlr23}. 

To simplify the notation, the following definitions are used,
\begin{align}
  \Delta_i(\fml{S};\fml{E},\cf) & \coloneq
  \left(\cf(\fml{S}\cup\{i\};\fml{E})-\cf(\fml{S};\fml{E})\right)
  \label{eq:def:delta}
  \\[2pt] 
  \varsigma(|\fml{S}|) & \coloneq
  \sfrac{|\fml{S}|!(|\fml{F}|-|\fml{S}|-1)!}{|\fml{F}|!} 
  \label{eq:def:vsigma}
\end{align}
(Observe that $\Delta_i$ is parameterized on $\fml{E}$ and $\cf$.)

Finally, let $\sv:\fml{F}\to\mbb{R}$, i.e.\ the theoretical SHAP score
for feature $i$, be defined by,\footnote{%
Throughout the paper, the definitions of $\Delta_i$ and $\sv$ are
explicitly associated with the characteristic function used in their
definition.}
\begin{equation} \label{eq:sv}
  \sv(i;\fml{E},\cf)\coloneq\sum\nolimits_{\fml{S}\subseteq(\fml{F}\setminus\{i\})}\varsigma(|\fml{S}|)\times\Delta_i(\fml{S};\fml{E},\cf) 
\end{equation}
In the case of Shapley values for XAI, the characteristic function
used in earlier work is $\cfn{e}$.
Given an instance $(\mbf{v},q)$, the SHAP score assigned to each
feature aims to measure the \emph{contribution} of that feature with
respect to the prediction. 
From earlier work, it is understood that a positive/negative value
indicates that the feature can contribute to changing the prediction,
whereas a value of 0 indicates no
contribution~\cite{kononenko-jmlr10}.

\begin{table}[t]
  \centering
  \scalebox{0.95}{\begin{tabular}{ccc}\toprule
  Acronym & Meaning & Definition \\
  \toprule
  $\shap{T}$ & Theoretical SHAP scores & Uses $\cfn{e}$
  (see~\eqref{eq:cfs}) in~\eqref{eq:sv}
  \\
  $\shap{E}$ & Estimated SHAP scores & Output of tool SHAP 
  \\
  \midrule[0.75pt]
  $\nshap{T}$ & Theoretical nuSHAP scores & Uses $\cfn{a}$
  (see~\eqref{eq:cfa}) in~\eqref{eq:sv}
  \\
  $\nshap{E}$ & Estimated nuSHAP scores & Output of tool nuSHAP 
  \\
  \bottomrule
\end{tabular}
}
  \caption{
    SHAP scores considered throughout the paper.} 
  \label{tab:shaps}
\end{table}

Throughout the paper, the acronyms from~\cref{tab:shaps} will be
used. The subscript $\mathrm{T}$ denotes a theoretical value, obtained
from the exact solution of~\eqref{eq:sv} using a specific
characteristic function.
Earlier work on XAI considered~$\cfn{e}$ (see~\eqref{eq:cfs}).
In contrast, the subscript $\mathrm{E}$ denotes an estimated value,
obtained from approximating the solution of~\eqref{eq:sv}. This
approximate value is most often obtained using the tool
SHAP~\cite{lundberg-nips17}.
This paper exploits a new characteristic function~\cite{lhms-aaai25},
but it also details a new algorithm for estimating a newly proposed
SHAP score.

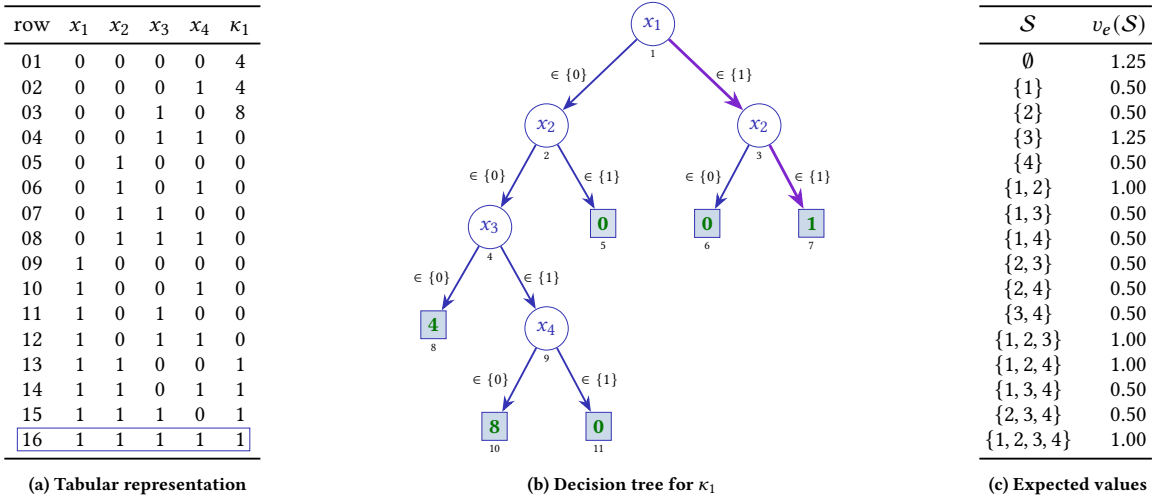
\begin{figure*}[t]
  \begin{subfigure}[b]{0.315\linewidth}
    \centering
    \renewcommand{\arraystretch}{0.9125}
    \renewcommand{\tabcolsep}{0.425em}
    \scalebox{0.95}{
      \begin{tabular}{cccccc} \toprule
  row & $x_1$ & $x_2$ & $x_3$ & $x_4$ & $\kappa_1$ \\
  \midrule[0.75pt]
  01 & 0 & 0 & 0 & 0 & 4 \\
  02 & 0 & 0 & 0 & 1 & 4 \\
  03 & 0 & 0 & 1 & 0 & 8 \\
  04 & 0 & 0 & 1 & 1 & 0 \\
  05 & 0 & 1 & 0 & 0 & 0 \\
  06 & 0 & 1 & 0 & 1 & 0 \\
  07 & 0 & 1 & 1 & 0 & 0 \\
  08 & 0 & 1 & 1 & 1 & 0 \\
  09 & 1 & 0 & 0 & 0 & 0 \\
  10 & 1 & 0 & 0 & 1 & 0 \\
  11 & 1 & 0 & 1 & 0 & 0 \\
  12 & 1 & 0 & 1 & 1 & 0 \\
  13 & 1 & 1 & 0 & 0 & 1 \\
  14 & 1 & 1 & 0 & 1 & 1 \\
  15 & 1 & 1 & 1 & 0 & 1 \\
  \tikzmarknode{e}{16} & 1 & 1 & 1 & 1 & \tikzmarknode{f}{1} \\
  \bottomrule
\end{tabular}
\begin{tikzpicture}[overlay,remember picture]
  \node[draw=midblue, thin, xshift=0.0pt, yshift=0.0pt, inner
    sep=1.5pt, fit=(e) (f)] {};
\end{tikzpicture}

    }
    \caption{Tabular representation} \label{rex01:tab:tr}
  \end{subfigure}
  \begin{subfigure}[b]{0.4\linewidth}
    \centering
    \scalebox{0.9}{
%
\forestset{
  BDT/.style={
    for tree={
      l=1.5cm,s sep=1.15cm,
      if n children=0{}{circle}, 
      draw=midblue,
      text=midblue,
      edge={
        my edge
      },
      edge=thick,
    }
  },
}
\begin{forest}
  BDT
  [{$x_1$}, label={[yshift=-6.875ex]{{\tiny1}}} 
    [{$x_2$}, label={[yshift=-6.875ex]{{\tiny2}}}, 
      edge label={node[midway,left,xshift=-1.5pt] {{\scriptsize$\in\{0\}$}}}
      [{$x_3$}, label={[yshift=-6.875ex]{{\tiny4}}}, 
        edge label={node[midway,left,xshift=-1.5pt] {{\scriptsize$\in\{0\}$}}}
        [\dghlight{\textbf{4}}, label={[yshift=-5.25ex]{{\tiny8}}},
          edge label={node[midway,left,xshift=-0.5pt]
            {{\scriptsize$\in\{0\}$}}}, rectangle, fill={tblue2!25} ]
        [{$x_4$}, label={[yshift=-6.875ex]{{\tiny9}}}, 
          edge label={node[midway,right,xshift=-1.5pt] {{\scriptsize$\in\{1\}$}}}
          [\dghlight{\textbf{8}}, label={[yshift=-5.25ex]{{\tiny10}}},
            edge label={node[midway,left,xshift=-0.5pt]
              {{\scriptsize$\in\{0\}$}}}, rectangle, fill={tblue2!25} ]
          [\dghlight{\textbf{0}}, label={[yshift=-5.25ex]{{\tiny11}}},
            edge label={node[midway,right,xshift=-0.575pt] {{\scriptsize$\in\{1\}$}}}, rectangle, fill={tblue2!25} ]
        ]
      ]
      [\dghlight{\textbf{0}}, label={[yshift=-5.25ex]{{\tiny5}}},
        edge label={node[midway,right,xshift=-0.5pt] {{\scriptsize$\in\{1\}$}}},
        rectangle, fill={tblue2!20} ]
    ]
    [{$x_2$}, label={[yshift=-6.875ex]{{\tiny3}}}, 
      edge={very thick,draw=purple3}, edge label={node[midway,right,xshift=0.25pt] {{\scriptsize$\in\{1\}$}}}
      [\dghlight{\textbf{0}}, label={[yshift=-5.25ex]{{\tiny6}}},
        edge label={node[midway,left,xshift=-0.5pt]
          {{\scriptsize$\in\{0\}$}}}, rectangle, fill={tblue2!25} ]
      [\dghlight{\textbf{1}}, label={[yshift=-5.25ex]{{\tiny7}}},
        edge={very thick,draw=purple3}, edge label={node[midway,right,xshift=-0.575pt] {{\scriptsize$\in\{1\}$}}}, rectangle, fill={tblue2!25} ]
    ]
  ]
\end{forest}
    }
    \caption{Decision tree for $\kappa_1$}  \label{rex01:fig:dt}
  \end{subfigure}
  %
  %
  %
  \begin{subfigure}{0.2575\linewidth}
    \centering
    \renewcommand{\arraystretch}{0.9125}
    \renewcommand{\tabcolsep}{0.225em}
    \scalebox{0.95}{
      \begin{tabular}{ccr} \toprule
  $\fml{S}$ & & $\cfn{e}(\fml{S})$ \\ \midrule[0.75pt] 
  $\emptyset$ & & 1.25 \\
  $\{1\}$ & & 0.50 \\
  $\{2\}$ & & 0.50  \\
  $\{3\}$ & & 1.25 \\
  $\{4\}$ & & 0.50 \\
  $\{1,2\}$ & & 1.00 \\
  $\{1,3\}$ & & 0.50 \\
  $\{1,4\}$ & & 0.50 \\
  $\{2,3\}$ & & 0.50 \\
  $\{2,4\}$ & & 0.50 \\
  $\{3,4\}$ & & 0.50 \\
  $\{1,2,3\}$ & & 1.00 \\
  $\{1,2,4\}$ & & 1.00 \\
  $\{1,3,4\}$ & & 0.50 \\
  $\{2,3,4\}$ & & 0.50 \\
  $\{1,2,3,4\}$ & & 1.00 \\
  \bottomrule
\end{tabular}

    }
    \caption{Expected values} \label{rex01:tab:avg}
  \end{subfigure}
  %
  %
  %
  \caption{
    ML model $\fml{M}_1$, adapted from Fig.~06(a)
    in~\cite{hms-ijar24}.
    As shown, the instance is $((1,1,1,1),1)$.
    For the DT, we have the set of AXps $\mbb{A}_1=\{\{1,2\}\}$ and
    the se of CXps $\mbb{C}_1=\{\{1\},\{2\}\}$. The expected values
    are used for computing the SHAP scores, as proposed
    in~\cite{lundberg-nips17}.}
  \label{fig:runex01}
\end{figure*}

\paragraph{Running examples.}
%
%
Two example ML models, $\fml{M}_1$ and $\fml{M}_2$ are used throughout
the paper. These are shown in~\cref{fig:runex01,fig:runex02}.

\begin{figure}[t]
  \begin{subfigure}[b]{0.55\linewidth}
    \centering
    \renewcommand{\tabcolsep}{0.325em}
    \scalebox{0.95}{
\begin{tabular}{cccccc} \toprule
  row & $x_1$ & $x_2$ & $\rho_2(\mbf{x})$ & $\alpha=\nfrac{1}{2}$ & $\alpha=\nfrac{1}{4}$
  \\ \toprule
  $1$ & $0$ & $0$ & $1-6\alpha$ & $-3$ & $-\nfrac{1}{2}$ \\
  $2$ & $0$ & $1$ & $1+2\alpha$ & $2$  & $\nfrac{3}{2}$  \\
  $3$ & $1$ & $0$ & $1$         & $1$  & $1$             \\
  \tikzmarknode{a}{$4$} & $1$ & $1$ & $1$ & $1$ & \tikzmarknode{b}{$\:1$}
  \begin{tikzpicture}[overlay,remember picture]%
    \node[draw=midblue, thin, xshift=-0.35pt, yshift=-0.25pt, inner  sep=2.0pt, fit=(a) (b)] {};%
  \end{tikzpicture}%
  \\
  \bottomrule%
\end{tabular}

    }
    \caption{Tabular representation (TR) of $\rho_2$}
    \label{rex02:tab:tr}
  \end{subfigure}
  ~
  \begin{subfigure}[b]{0.435\linewidth}
    \centering
    \renewcommand{\tabcolsep}{0.325em}
    \scalebox{0.95}{
\begin{tabular}{ccc} \toprule
  $\fml{S}$ & $\msf{rows}(\fml{S})$ & $\cfn{e}(\fml{S})$
  \\ \toprule
  $\emptyset$ & $1,2,3,4$ & $1-\alpha$ \\
  $\{1\}$ & $3,4$ & $1$ \\
  $\{2\}$ & $2,4$ & $1+\alpha$ \\
  $\{1,2\}$ & $4$ & $1$
  \\ \bottomrule
\end{tabular}

    }
    \caption{Expected values of $\rho_2$}
    \label{rex02:tab:avg}
  \end{subfigure}
  \caption{Simple ML model $\fml{M}_2$, with instance $((1,1),1)$, and
    $\alpha\not=0$.
    The expected values are computed for all possible sets of
    features. Clearly, $\mbb{A}_2=\mbb{C}_2=\{\{1\}\}$. }
  \label{fig:runex02}
\end{figure}

\section{The Flaws of $\shap{T}$ Scores} 

The fact that theoretical SHAP (i.e.\ $\shap{T}$) scores use the
expected value of an ML model raises a number of critical issues.
First, in the case of classification, where classes are categorical,
the expected value is ill-defined. 
Unfortunately, the use of expected values is more problematic. This
section overviews the known flaws of $\shap{T}$ scores, and briefly
discusses generalizations of those flaws to the case of regression
problems.


\subsection{Known Flaws of $\shap{T}$ Scores}

%
It is reasonably simple to show that $\shap{T}$ scores can be
misleading.
The key insight is that one can create examples of ML models where
feature influence for a prediction is self-evident, and then force the
computation of the $\shap{T}$ scores to produce misleading values by
assigning no importance to influent features, and by assigning some
importance to non-influent features. This basic insight can then be
used to identify different issues where misleading information is
clearly unsatisfactory.
As a consequence, the tool SHAP~\cite{lundberg-nips17} aims to
approximate values that can be misleading.

Although these observations have gone unnoticed in the many thousands
of publications that build on or exploit SHAP scores, simple examples
serve to illustrate what the critical limitation is. Since $\shap{T}$
scores use a characteristic function that computes expected values,
then one can use regions of the feature space, that are
non-interesting in terms of the prediction, to influence the
computation of the expected values. Then, 
one is capable of modifying the theoretical SHAP scores as one wishes,
thereby destroying any correlation that $\shap{T}$ scores might have
with respect to real feature influence. 

\begin{figure}[t]
  \begin{subfigure}[b]{0.615\linewidth}
    \scalebox{0.885}{
      \renewcommand{\tabcolsep}{0.225em}
      \begin{tabular}{cccccc} \toprule
         & $\svn{e}(1)$ & $\svn{e}(2)$ & $\svn{e}(3)$ &
        $\svn{e}(4)$ & Rank
        \\
        \toprule
        $\kappa_1$ & 0.000 & 0.000 & -0.042 & -0.208 &
        $\langle4,3,1{:}2\rangle$ \\
        \bottomrule
      \end{tabular}
    }
    \caption{Scores for $\fml{E}_1$}
    \label{rex01:tab:svs}
  \end{subfigure}
  {~~}
  \begin{subfigure}[b]{0.35\linewidth}
    \scalebox{0.885}{
      \renewcommand{\tabcolsep}{0.225em}
      \begin{tabular}{cccccc} \toprule
         & $\svn{e}(1)$ & $\svn{e}(2)$ & Rank
        \\
        \toprule
        $\tau_2$ & 0 & $\alpha$ & $\langle2,1\rangle$ \\
        \bottomrule
      \end{tabular}
    }
    \caption{Scores for $\fml{E}_2$}
    \label{rex02:tab:svs}
  \end{subfigure}
  \caption{$\shap{T}$ scores for $\fml{E}_1$ and
    $\fml{E}_2$. These are the values that the tool
    SHAP~\cite{lundberg-nips17} approximates.}
  \label{fig:svs}
\end{figure}

\begin{example}
  For the two running examples, with explanation problems $\fml{E}_1$
  and $\fml{E}_2$, the computed $\shap{T}$ scores are shown
  in~\cref{fig:svs}.
  These are the values that the tool SHAP~\cite{lundberg-nips17}
  approximates.
  Even if the tool SHAP were successful in approximating the values
  shown in~\cref{fig:svs}, that would be useless. It should be evident
  that the $\shap{T}$ scores computed using $\cfn{e}$
  are unsatisfactory, since the only influent features are assigned 
  \emph{no} importance. 
\end{example}

\begin{example}
  The arguments above can be clarified with the DT
  of~\cref{rex01:fig:dt}. The values given to the terminal nodes 8, 10
  and 11 are chosen with the purpose of fixing the $\shap{T}$ scores
  of features 1 and 2 to 0, and the $\shap{T}$ scores of features 3
  and 4 to a value other than 0. The approach to compute these values
  is simple, and described in detail in earlier 
  publications~\cite{msh-cacm24,hms-ijar24}.
\end{example}

\subsection{$\shap{T}$ Scores Flawed Beyond Classification} 
\label{sec:issues}

%
A possible criticism of our earlier work~\cite{hms-ijar24,msh-cacm24}
is that the use of classifiers is somewhat artificial, since we impose
that classes be viewed as ordinal values. The use of some regression
models, e.g.\ regression trees, can also be criticized, since the
classifier simply computes a constant number of fixed values.
Nevertheless, one can devise regression models for which a
non-countable number of values can be predicted, and for which the
$\shap{T}$ scores are again misleading. 

\begin{example}(Regression model $\fml{M}_3$.) \label{ex:rm02}
  We consider a regression problem defined over two real-valued
  features, taking values from interval
  $[-\sfrac{1}{2},\sfrac{3}{2}]$.
  Thus, we have $\fml{F}=\{1,2\}$,
  $\mbb{D}_1=\mbb{D}_2=\mbb{D}=[-\sfrac{1}{2},\sfrac{3}{2}]$,
  $\mbb{F}=\mbb{D}\times\mbb{D}$.
  (We also let $\mbb{D}^{+}=[\sfrac{1}{2},\sfrac{3}{2}]$.)
  %
  In addition, the regression model maps to real values,
  i.e.\ $\mbb{V}=\mbb{R}$, and is defined as follows:
  \[
    \rho_3(x_1,x_2) =
    \left\{
    \begin{array}{lcl}
      x_1 & ~~ & \tn{if $x_1\in\mbb{D}^{+}$} \\ 
      x_2-2 & ~~ & \tn{if
        $x_1\not\in\mbb{D}^{+}\land{x_2}\not\in\mbb{D}^{+}$}\\
      x_2+1 & ~~ & \tn{if
        $x_1\not\in\mbb{D}^{+}\land{x_2}\in\mbb{D}^{+}$}\\
    \end{array}
    \right.
  \]
  As a result, the regression model is represented by
  $\fml{M}_3=(\fml{F},\mbb{F},\mbb{V},\rho_3)$.
  Moreover, we assume the target instance to be
  $(\mbf{v}_3,q_3)=((1,1),1)$, and so the explanation problem becomes 
  $\fml{E}_3=(\fml{M}_3,(\mbf{v}_3,q_3))$.
\end{example}


\section{New SHAP Scores \& Implementation}
\label{sec:nushap}

\subsection{Selecting the Characteristic Function} \label{ssec:cf}

Recent work dissected the causes for the flaws of the theoretical
SHAP scores (i.e.\ $\shap{T}$)~\cite{lhms-aaai25,lhams-corr24}. 
Concretely, these flaws 
result solely from
the choice of characteristic function. 
This characteristic function is used for example in the SHAP
tool~\cite{lundberg-nips17}, but was studied in earlier
works \cite{kononenko-jmlr10,kononenko-kis14}. More importantly, this
recent work \cite{lhms-aaai25,lhams-corr24} proposed  
alternative characteristic functions that eliminate all of the issues
reported in earlier works \cite{msh-cacm24,hms-ijar24}.
%
%
Accordingly,
in this paper we consider the characteristic function
$\cfn{a}:2^{\fml{F}}\to\mbb{R}$, defined as follows:
\begin{equation} \label{eq:cfa}
  \begin{array}{lcl}
    \cfn{a}(\fml{S}) & \coloneq &
    \left\{
    \begin{array}{lcl}
      1 & \quad & \tn{if $\waxp(\fml{S})$} \\[3pt]
      0 & \quad & \tn{otherwise} \\
    \end{array}
    \right.
  \end{array}
\end{equation}
%

The novel characteristic function is inspired by those commonly used
in game theory~\cite{shapley-apsr54}. The motivation then and now is
to assign importance to the elements (features or voters) which are
\emph{critical} for changing the value of a decision of interest.
%
In the case of voting power, one assigns importance to voters that
cause coalitions to become winning when the voter is included, and
that are losing when the voter is not included.
In the case of explainability, one assigns importance to a feature
that causes fixed sets of features to be sufficient for the prediction
when the feature is also fixed, and that causes fixed sets of features
not to be sufficient for the prediction when the feature is not fixed.

It is clear that, given the novel characteristic function $\cfn{a}$,
we can use the definition of Shapley values to obtain SHAP scores
different from those obtained with $\cfn{e}$.
As noted earlier, the new SHAP scores will be referred to by the
acronym $\nshap{T}$;
their properties
are the topic of recent work~\cite{lhms-aaai25,lhams-corr24}. 

It is also apparent that we impose few constraints on the 
ML models for which the novel SHAP scores can be computed. We
\emph{just} need to be able to decide whether a set of selected (and
so fixed) features is sufficient for predicting the value we are
interested in.
%
%
It should also be
underscored that the similarity predicate allows us to consider both
classification and regression models.
%
%
%
Finally, as underscored in recent work~\cite{lhms-aaai25}, the
proposed characteristic function reveals a fundamental relationship
between explanations based on feature selection, and those based on
feature attribution.
The following result will be used in later sections, and it is a
corollary of~\cite[Proposition~3]{hims-aaai23}.

\begin{proposition} \label{prop:nochange}
  Let $i\in\fml{F}$ be an irrelevant feature. Then,
  \[
  \forall(\fml{S}\subseteq(\fml{F}\setminus\{i\})).
  \left[\cfn{a}(\fml{S}\cup\{i\})=\cfn{a}(\fml{S})\right]
  \]
\end{proposition}



From the definition, it is clear that $\cfn{a}$ is monotonically
increasing. Also, the following result is immediate.
\begin{proposition}
  $\Delta_i(\fml{S})\in\{0,1\}$ (see~\cref{eq:def:delta}),
  where $\fml{S}\subseteq\fml{F}\setminus\{i\}$ and $i\in\fml{F}$. 
\end{proposition}

Furthermore,
%
a feature $i\in\fml{F}$ is \emph{critical} for
$\fml{S}\subseteq\fml{F}$ if $\cfn{a}(\fml{S})=0$ and
$\cfn{a}(\fml{S}\cup\{i\})=1$.%
\footnote{We adopt the concept of critical elements from game theory,
that can be traced at least to the work of Shapley\&Shubik on voting
power~\cite{shapley-apsr54}.}
Thus, we have the predicate, 
\begin{equation} \label{eq:critic}
  \critic(i,\fml{S}) ~~ \coloneq ~~ \left(\cfn{a}(\fml{S}\cup\{i\})=1\right)\land\left(\cfn{a}(\fml{S})=0\right)
\end{equation}
Clearly, $\critic(i,\fml{S})$ holds iff $\Delta_i(\fml{S})$ holds
(see~\eqref{eq:def:delta}).
In terms of logic-based explanations, a feature $i\in\fml{F}$ is
critical for a set $\fml{S}\subseteq\fml{F}\setminus\{i\}$ of fixed
features (i.e.\ features taking the values dictated by $\mbf{v}$) if
$\fml{S}$ is not a WAXp and $\fml{S}\cup\{i\}$ is a WAXp.

\begin{algorithm}[t]
  \begin{flushleft}
  \hspace*{\algorithmicindent}
  \textbf{Input}: {%
    $\fml{G}=(N,\cf)$: game; $\epsilon$: error; $\alpha$: prob.~error exceeded
  }\\
  \hspace*{\algorithmicindent}
  \textbf{Output}: {$[\she{1},\ldots,\she{n}]$}\\
\end{flushleft}

\begin{algorithmic}[1]
  \Function{$\cgtalg$}{$\fml{G}=(N,\cf)$, $\epsilon$, $\alpha$}
  \State{Estimate value of $r$ given $\epsilon$, $\alpha$, $\cf$, $N$}
  \State{$\she{i}\gets0$ for $i\in\{1,\ldots,n\}$}
  \For{$\iter\gets{r}$ \kwdownto 1}
  \State{Pick $\perm\in\psi(N)$ with uniform probability}
  \For{$i \in \{1,\ldots,n\}$}
  \State{$\pref(i)\gets\{\fml{P}(1),\ldots,\fml{P}(k-1)\,|\,\mathrm{~s.t.~}i=\fml{P}(k)\}$}
  \State{$\Delta_i(\pref(i))\gets\cf(\pref(i)\cup\{i\})-\cf(\pref(i))$}
  \State{$\she{i}\gets\she{i}+\Delta_i(\pref(i))$}
  \EndFor
  \EndFor
  \State{$\she{i}\gets\sfrac{\she{i}}{r}$ for $i\in\{1,\ldots,n\}$}
  \State{\Return $[\she{1},\ldots,\she{n}]$}
  \EndFunction
\end{algorithmic}

  \caption{Shapley value estimation~\cite{tejada-cor09}}
  \label{alg:cgt}
\end{algorithm}

\subsection{Rigorous Estimation of Shapley Values} \label{ssec:svs}

The complexity of rigorously computing Shapley
values is well-known to be
unwieldy~\cite{vandenbroeck-jair22,barcelo-jmlr23}. As a result,
except for small examples or special tractable cases, the most common
solution is to approximate the Shapley value.
In contrast with earlier works on applying Shapley values in XAI, we
opt to estimate the Shapley value using an algorithm known for its
strong theoretical guarantees~\cite{tejada-cor09}. (This algorithm
will be referred to as CGT, and is shown as~\cref{alg:cgt}.)
The inputs to the algorithm consist of a game~\cite{elkind-bk12},
characterized by a number of elements 
and a characteristic function $\cf:2^N\to\mbb{R}$, such that
$\cf(\emptyset)=0$, and also the values of $\epsilon$ and $\alpha$.
$\epsilon$ denotes the error of Shapley value estimation, that one
wants to guarantee with probability $1-\alpha$. The algorithm
executes a number $r$ of runs of a basic estimation procedure, such
that $\prob(|\she{i}-\shv{i}|\le\epsilon)\ge1-\alpha$, where $\shv{i}$
denotes the exact Shapley value for feature $i$, and $\she{i}$ denotes
the estimate of the Shapley value. $r$ denotes the number of times the
main loop of the CGT algorithm is executed; its estimation is detailed
in~\cite{tejada-cor09}.
The algorithm iteratively picks a permutation of $N$
from the set of permutations $\psi(N)$. Each permutation is
then used for defining a total of $n$ sets of features, each with a
different size, for each of which the value of $\Delta_i$ is computed
and used for updating $\she{}$.
\begin{example} \label{ex:perm}
  Let the permutation be $\fml{P}=\langle3,1,4,2\rangle$.
  Thus, from~\cref{alg:cgt}, the prefixes $\pref$ are:
  \[
  \begin{array}{lcl}
    \pref(1) = \{3\} & \quad & \pref(3) = \emptyset \\
    \pref(2) = \{3,1,4\} & \quad & \pref(4) = \{3,1\} \\
  \end{array}
  \]
\end{example}
By inspection, it is clear that~\cref{alg:cgt} runs in polynomial
time~\cite{tejada-cor09}, if $\cf$ is also evaluated in
polynomial-time.
Also, several low-level optimizations can be envisioned; these are
beyond the scope of this paper.
Furthermore, two observations are in order.
A first observation is that, given $\cfn{a}$, we have in fact a simple
game, because $\cfn{a}$ is monotonically increasing. 
A second key observation is that, because of~\cref{prop:nochange},
$\Delta_i$ is never incremented by~\cref{alg:cgt} when $i$ is
irrelevant. As a result, the SHAP score for irrelevant features will
\emph{always} be 0, i.e.\ if $i$ is irrelevant, then
$\she{i}=\shv{i}=0$. %
(This may also happen with relevant features.)
%
Thus, 
the issues with $\shap{T}$ scores reported in earlier
work~\cite{msh-cacm24} \emph{cannot} 
occur if the characteristic function is $\cfn{a}$ and the estimate
is computed with CGT.

\subsection{Deciding Weak Abductive Explanations} \label{ssec:waxp}

A key component of estimating rigorous SHAP scores is the ability to
efficiently decide whether a set of features is a weak AXp.
%
Clearly,
logic-based computation of explanations can be used in this
context~\cite{ms-rw22,ms-iceccs23,ms-isola24}.
However, for large-scale datasets and respective 
ML models,
this approach is unlikely to scale.
First, there is the complexity of reasoning, which is still a
challenge for highly complex ML models despite recently
observed progress~\cite{ihmpims-kr24}.
Second, even if deciding whether a set of features is a weak AXp could
be done reasonably efficiently, the very large number of tests that
must be considered for estimating Shapley values would 
likely make the running times prohibitively large.
The consequence is that deciding whether a set of features is a weak
AXp should be attained with a negligible running time, ideally
with polynomial-time guarantees.

Therefore, instead of exploiting a rigorous model-based approach for 
explaining an ML model, we consider instead a rigorous
\emph{data-based} approach.
Concretely, we propose instead to exploit \emph{sample-based
explanations}~\cite{cooper-ecai23} (sbXps). sbXps are computed with
respect to a \emph{sample} of the feature space, i.e.\ a set of
instances, and are rigorous with respect to that sample. For example,
the sample might be the original dataset, or consist of the sampling
carried out by tools such as LIME~\cite{guestrin-kdd16},
SHAP~\cite{lundberg-nips17} or Anchor~\cite{guestrin-aaai18}, or an
aggregation of both. However, sbXps are \emph{rigorous} given the
sample. Thus, in cases where the sample is the feature space,
sbXps match AXps/CXps.

A \emph{sample space} $\mbb{S}$ is a subset of feature space,
i.e.\ $\mbb{S}\subseteq\mbb{F}$. (We refer to a sample space as a
\emph{dataset} when the prediction is known for each point in the
sample space.)
A set of features $\fml{X}\subseteq\fml{F}$ is a weak sbAXp if,
\begin{equation} \label{def:wsbaxp}
  \forall(\mbf{x}\in\mbb{S}).\left(\mbf{x}_{\fml{X}}\simrelop\mbf{v}_{\fml{X}}\right)\limply(\pi(\mbf{x})\simrelop\pi(\mbf{v}))
\end{equation}
If no proper subset of $\fml{X}$ is also a weak sbAXp, then $\fml{X}$ is
declared an sbAXp.
Similarly, a set of features $\fml{Y}\subseteq\fml{F}$ is a weak sbCXp
if,
\begin{equation} \label{def:wsbcxp}
  \exists(\mbf{x}\in\mbb{S}).\left(\mbf{x}_{\fml{F}\setminus\fml{Y}}\simrelop\mbf{v}_{\fml{F}\setminus\fml{Y}}\right)\land(\pi(\mbf{x})\not\simrelop\pi(\mbf{v}))
\end{equation}
If no proper subset of $\fml{Y}$ is a weak sbCXp, then $\fml{Y}$ is
declared an sbCXp.%
\footnote{As with earlier work on logic-based
XAI~\cite{ms-rw22,ms-isola24}, it is simple to prove 
minimal hitting set duality between abductive and contrastive
explanations.}

\begin{table}[t]
  \centering
  \renewcommand{\arraystretch}{0.90}
  \renewcommand{\tabcolsep}{0.5em}
  \scalebox{0.925}{
    \begin{tabular}{cccccc} \toprule
  $\fml{S}$ & sbWCXp & sbCXp & sbWAXp & sbAXp & $\{i\,|\,\critic(i,\fml{S})\}$
  \\
  \midrule[0.75pt]
  $\emptyset$ & \xmark & \xmark & \xmark & \xmark & $\emptyset$ \\
  $\{1\}$     & \cmark & \cmark & \xmark & \xmark & $\{2\}$ \\
  $\{2\}$     & \cmark & \cmark & \xmark & \xmark & $\{1\}$ \\
  $\{3\}$     & \xmark & \xmark & \xmark & \xmark & $\emptyset$ \\
  $\{4\}$     & \xmark & \xmark & \xmark & \xmark & $\emptyset$ \\
  $\{1,2\}$   & \cmark & \xmark & \cmark & \cmark & $\emptyset$ \\
  $\{1,3\}$   & \cmark & \xmark & \xmark & \xmark & $\{2\}$ \\
  $\{1,4\}$   & \cmark & \xmark & \xmark & \xmark & $\{2\}$ \\
  $\{2,3\}$   & \cmark & \xmark & \xmark & \xmark & $\{1\}$ \\
  $\{2,4\}$   & \cmark & \xmark & \xmark & \xmark & $\{1\}$ \\
  $\{3,4\}$   & \xmark & \xmark & \xmark & \xmark & $\emptyset$ \\
  $\{1,2,3\}$ & \cmark & \xmark & \cmark & \xmark & $\emptyset$ \\
  $\{1,2,4\}$ & \cmark & \xmark & \cmark & \xmark & $\emptyset$ \\
  $\{1,3,4\}$ & \cmark & \xmark & \xmark & \xmark & $\{2\}$ \\
  $\{2,3,4\}$ & \cmark & \xmark & \xmark & \xmark & $\{1\}$ \\
  $\{1,2,3,4\}$ & \cmark & \xmark & \cmark & \xmark & $\emptyset$ \\
  \bottomrule
\end{tabular}

  }
  \caption{Computation of sb(W)CXps, sb(W)AXps, and critical elements
    for the dataset in~\cref{rex01:tab:tr}. For sb(W)CXps, each picked
    feature is allowed to take any value from its domain. For
    sb(W)AXps, each picked feature is fixed to the value dictated by
    $\mbf{v}$.}
  \label{tab:xpscrit}
\end{table}

\begin{example} \label{ex:xpscrit}
  \cref{tab:xpscrit} summarizes the sbWCXps/\-sbCXps,
  sbWAXps/\-sbAXps, and the critical elements (for the computation of
  $\nshap{T}$ scores) for the example dataset shown in~\cref{rex01:tab:tr}.
  The computation of explanations is based on the definitions above.
\end{example}

Given a sample space, there exists a polynomial-time algorithm for
computing an sbAXp, and so for deciding whether a set of features is
an sbWAXp~\cite{cooper-ecai23}. (For the case of complete truth
tables, polynomial-time algorithms are also described
in~\cite{hms-corr23a}.) 
It is straightforward to devise a polynomial-time algorithm for
deciding whether a set of features $\fml{W}$ is an sbWAXp. One simple
algorithm is outlined in~\cref{alg:sbwaxp}.
\begin{algorithm}[t]
  \begin{flushleft}
  \hspace*{\algorithmicindent}
  \textbf{Input}: {$\msf{DSet}$: Dataset; $(\mbf{v},c)$: Instance; $\fml{W}$: Tentative sbWAXp}\\
  \hspace*{\algorithmicindent}
  \textbf{Output}: {\True~ if $\fml{W}$ is sbWAXp; \False~ otherwise}\\
\end{flushleft}

\begin{algorithmic}[1]
  \Function{$\sbwaxp$}{$\msf{DSet}$, $(\mbf{v},c)$, $\fml{W}$}
  \For{$i\in\msf{DSet}.\msf{rows}()$}
  \State{$\msf{covered}(i)\gets\False$}
  \EndFor
  \For{$j\in\fml{W}$}
  \For{$i\in\msf{DSet}.\msf{rows}()$}
  \If{$\msf{DSet}.\msf{pred}(i)\not=c$ \kwand $\msf{DSet}.\msf{val}(i,j)\not=v_j$}
  \State{$\msf{covered}(i)\gets\True$}
  \EndIf
  \EndFor
  \EndFor
  \For{$i\in\msf{DSet}.\msf{rows}()$}
  \If{$\msf{DSet}.\msf{pred}(i)\not=c$ \kwand\kwnot $\msf{covered}(i)$}
  \State{\Return \False}
  \Comment{$\fml{W}$ is not a sbWAXp}
  \EndIf
  \EndFor
  \State{\Return \True}
  \Comment{$\fml{W}$ is a sbWAXp}
  \EndFunction
\end{algorithmic}

  \caption{Deciding whether set of features is an sbWAXp}
  \label{alg:sbwaxp}
\end{algorithm}
If there exists an individual sample $(\mbf{u},d)$ in the dataset,
with $d\not=c$, such that
$\mbf{u}_{\fml{W}}\simrelop\mbf{v}_{\fml{W}}$, then $\fml{W}$ is
not an sbWAXp; otherwise it is.
The features in $\fml{W}$ are analyzed one at a time, and this is
justified by efficiency in practice.
Clearly, for $n$ samples, the running time is $\fml{O}(mn)$. (Observe
that the complexity of $\fml{O}(mn)$ for deciding whether a set of
features is an sbWAXp is more efficient than the one proposed in%
~\cite{cooper-ecai23} for computing one sbAXp; however, the problems
being solved are different.)
The complexity can be refined as follows.
If the set of sbCXps $\mbb{C}$ is pre-computed, and the size of the 
candidate sbWAXp is $|\fml{W}|$, then the running time is
$\fml{O}(|\mbb{C}|\times|\fml{W}|)$.

\begin{table}[t]
  \renewcommand{\tabcolsep}{0.775em}
  \scalebox{0.925}{
    \begin{tabular}{cc} \toprule
      $j$ & Rows that become covered due to $j$
      \\ \midrule[0.75pt]
      1 & $\{01, 02, 03, 04, 05, 06, 07, 08\}$ \\
      3 & $\{09, 10\}$ \\
      4 & $\{11\}$ \\
      \bottomrule
    \end{tabular}
  }
  \caption{Execution of \cref{alg:sbwaxp} with $\fml{W}=\{1,3,4\}$,
    using the dataset from~\cref{rex01:tab:tr}.}
  \label{tab:issbwaxp}
\end{table}

\begin{example}
  We illustrate the execution~\cref{alg:sbwaxp} using the tabular
  representation of the running example in~\cref{rex01:tab:tr} as the
  dataset.
  \cref{tab:issbwaxp} summarizes the main steps of the algorithm's
  execution. We assume that features are analyzed in order. If feature
  1 is fixed, then the rows in set $\{01,02,03,04,05,06,07,08\}$ are
  covered, because these rows would require feature 1 to take a
  different value for consistency.
  Afterwards, if feature 3 is also fixed, then the rows in set
  $\{09,10\}$ are covered. (Clearly, rows with the same prediction are
  non-interesting.) Finally, if feature 4 is also fixed, then row
  11 is covered. Overall, among the rows yielding a prediction other
  than 1, row $12$ is not covered. Thus, $\{1,3,4\}$ is not an sbWAXp.
\end{example}

\begin{example}
  Given the running example, the permutation and the prefixes
  from~\cref{ex:perm}, and the results in~\cref{tab:xpscrit},
  we can now compute the $\Delta_i(\pref(i))$, as follows,
  \[
  \begin{array}{lcl}
    \pref(1) = \{3\} & ~~ & \Delta_1(\pref(1))=\critic(1,\{3\})=0 \\
    \pref(2) = \{3,1,4\} & ~~ & \Delta_2(\pref(2))=\critic(2,\{1,3,4\})=1 \\
    \pref(3) = \emptyset & ~~ & \Delta_3(\pref(3))=\critic(3,\emptyset)=0 \\
    \pref(4) = \{3,1\} & ~~ & \Delta_4(\pref(4))=\critic(4,\{1,3\})=0 \\
  \end{array}
  \]
  Moreover, each contribution for element $i$ serves to update the
  value of $\she{i}$, as shown in~\cref{alg:cgt}.
\end{example}

As noted earlier, 
sample-based explanations are data-accurate, but not model-accurate.
One advantage is that explanations are model-agnostic, i.e.\ details
of the model need not be known. In contrast, one disadvantage is that
computed explanations, depending on the quality of the sample, may not
reflect entirely the explanations obtained from the ML model.
Nevertheless, the experimental set up ensures a rigorous and fair
comparison with the
tool SHAP.


\section{Experimental Evidence}
\label{sec:res}

\paragraph{nuSHAP \& experimental procedure.}
%
nuSHAP is a novel prototype explainer by feature attribution, that
implements the ideas detailed in the previous section.
%
%
Shapley values are estimated with the well-known CGT
algorithm~\cite{tejada-cor09} (see~\cref{alg:cgt}).
The characteristic function used is $\cfn{a}$~\cite{lhms-aaai25}.
%
For scalability, instead of employing standard model-based
explanations, the characteristic function $\cfn{a}$ is defined in
terms of sample-based explanations, building on recent
work~\cite{cooper-ecai23}. Sample-based weak AXps are decided
with~\cref{alg:sbwaxp}.
%
%
The experiments are organized into (i) analysis of boolean functions;
and (ii) direct comparison with the existing SHAP tool.
%
%
%
All experiments were run on a MacBook Pro with a 6-Core Intel
Core i7 2.6 GHz processor with 16 GByte RAM, running macOS Sonoma.

\paragraph{Boolean functions.}
%
As underscored earlier in the paper, the CGT algorithm together with
$\cfn{a}$ guarantees that irrelevant features are \emph{never}
assigned an estimate of their Shapley value other than 0. Therefore,
\emph{any} of the issues studied in earlier
work~\cite{msh-cacm24,hms-ijar24} \emph{do not} occur.

\paragraph{Tabular \& image data -- SHAP samples.}
%
%
%
%
%
The second experiment aims at assessing the quality of the SHAP
tool~\cite{lundberg-nips17} at ranking features in terms of their
importance for a given prediction.
To ensure a fair comparison, the sampling performed by the tool SHAP
was recorded. This sampling was then used by nuSHAP
for computing the $\nshap{E}$ scores.

The tools SHAP\footnote{Available from~\url{https://github.com/slundberg/shap}.}
and nuSHAP were assessed on several well-known
classifiers~\cite{zhou-bk21}, namely: 
logistic regression (LR), 
decision tree (DT), 
$k$-nearest neighbors ($k$NN) classifier, 
boosted trees (BT), 
and Convolutional Neural Network (CNN).
LR, DT, and $k$NN models are trained using
scikit-learn~\cite{pedregosa2011scikit},
BT models are trained using the XGBoost algorithm~\cite{chen2016xgboost},
while CNN models are trained using TensorFlow\footnote{\url{https://www.tensorflow.org/}.}.
The comparison was conducted across a range of widely used tabular 
classification datasets selected from the PMLB
benchmark~\cite{Olson2017PMLB},
as well as the MNIST dataset~\cite{deng2012mnist} of handwritten
digits (0–9).
The selected datasets are divided into five sets.%
\footnote{%
The numbers of features and classes are shown in parentheses.
}
The first set comprises \emph{adult} (14, 2) %
and \emph{corral} (6, 2),
the second set comprises \emph{iris} (4, 3) and \emph{mux6} (6, 2)
the third set comprises \emph{connect\_4} (42, 3), \emph{spambase} (57, 2) and \emph{spectf} (44, 2),
and the fourth set comprises \emph{clean1} (166, 2), \emph{coil2000}
(85, 2), and \emph{dna} (180, 3).
The last set comprises \emph{MNIST} ($28 \times 28$, 10).
For each tabular dataset, we randomly picked 50 tested instances for computing
nuSHAP and SHAP scores.
For the MNIST dataset, we randomly selected 20 test instances to compute these scores.
Moreover, we chose different model-agnostic SHAP explainers when
computing SHAP scores.
\emph{ExactExplainer} was used for the first and second sets,
\emph{PermutationExplainer} was used for the third set,
and \emph{SamplingExplainer} was used for the fourth and last set.
All explainers are provided with the entire training data so that the
SHAP tool can draw samples from it.
However, for MNIST a reduced training data was used, to curb the size
of sampling.
For the nuSHAP tool, the parameters used for computing the $\nshap{E}$
scores were $\epsilon=0.0015$ and $\alpha=0.015$, for all the
tested instances.
%

\begin{table*}[t] 
  \hspace*{-0.25cm}
  \renewcommand{\tabcolsep}{0.415em}
  \scalebox{0.925}{
    \begin{tabular}{llrrrrrrrrrrr}
      \toprule
      &           & adult & corral & iris & mux6 & connect\_4 & spambase & spectf & clean1 & coil2000 & dna  & MNIST \\
      \midrule
      \multirow{2}{*}{Min}  & nuSHAP vs.~SHAP      & 0.08  & 0.17   & 0.31 & 0.32 & 0.0       & 0.01     & 0.0    & 0.0    & 0.0      & 0.0  & 0.0 \\
      & nuSHAP vs.~abs(SHAP) & 0.05  & 0.12   & 0.27 & 0.32 & 0.0       & 0.05     & 0.0    & 0.0    & 0.0      & 0.03 & 0.0 \\
      \midrule
      \multirow{2}{*}{Max}  & nuSHAP vs.~SHAP      & 0.96  & 0.96   & 0.94 & 0.97 & 0.9       & 0.94     & 0.91   & 0.69   & 0.69     & 0.88 & 0.06 \\
      & nuSHAP vs.~abs(SHAP) & 0.88  & 0.97   & 0.94 & 0.95 & 0.77      & 0.94     & 0.91   & 0.88   & 0.69     & 0.88 & 0.06 \\
      \midrule
      \multirow{2}{*}{Mean} & nuSHAP vs.~SHAP      & 0.37  & 0.53   & 0.84 & 0.7  & 0.21      & 0.41     & 0.2    & 0.12   & 0.05     & 0.17 & 0.0 \\
      & nuSHAP vs.~abs(SHAP) & 0.31  & 0.5    & 0.84 & 0.69 & 0.19      & 0.42     & 0.19   & 0.17   & 0.08     & 0.43 & 0.0 \\
      \bottomrule
    \end{tabular}
  }
  \caption{Summary of RBO values for all the tested instances, including the minimum, maximum, and mean RBO values.}
  \label{tab:rbo_sum}
\end{table*}

\begin{table*}[t] 
  \centering
  \scalebox{0.95}{
    \begin{tabular}{lrrrrrrrrrrr}
      \toprule
      & adult & corral & iris & mux6 & connect\_4 & spambase & spectf & clean1 & coil2000 & dna & MNIST \\
      \midrule
      SHAP & 3.4 & \textbf{0.1} & \textbf{0.0} & \textbf{0.0} & 21.7 & \textbf{0.5} & \textbf{0.7} & 6.8 & 28.2 & 23.0 & 281.3 \\
      nuSHAP & \textbf{1.9} & 1.5 & 1.5 & 1.5 & \textbf{4.5} & 2.7 & 2.9 & \textbf{2.7} & \textbf{1.7} & \textbf{4.5} & \textbf{48.9} \\
      \#Samples & 68045.5 & 63.7 & 955.4 & 63.9 & 9202.5 & 13654.1 & 36459.5 & 3929.6 & 2960.8 & 2756.8 & 2929.9 \\
      \bottomrule
    \end{tabular}
  }
  \caption{Average runtime (in seconds) for computing SHAP and nuSHAP scores and average \#samples used for computing nuSHAP scores.}
  \label{tab:time_tabular}
\end{table*}

\begin{figure*}
  \hspace*{-0.75cm}
  \begin{subfigure}[b]{0.475\textwidth}
    \centering
    \includegraphics[width=1.0575\linewidth]{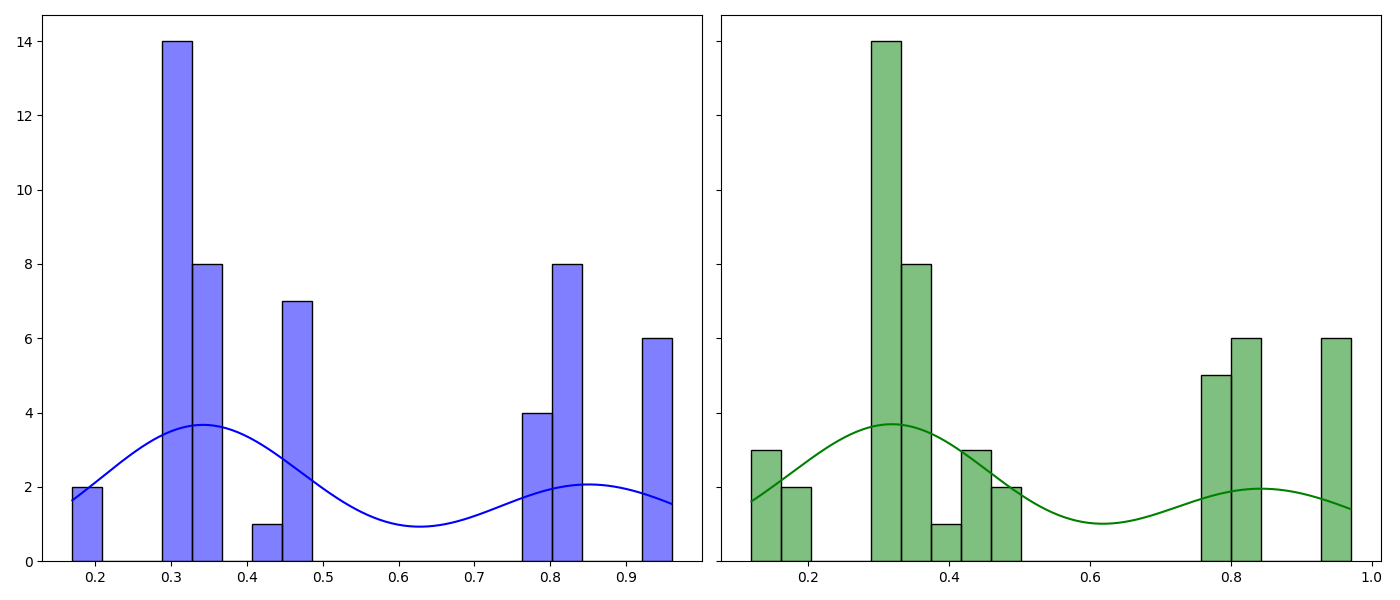}
    \caption{corral}
  \end{subfigure}
  \quad
  \begin{subfigure}[b]{0.475\textwidth}
    \centering
    \includegraphics[width=1.0575\linewidth]{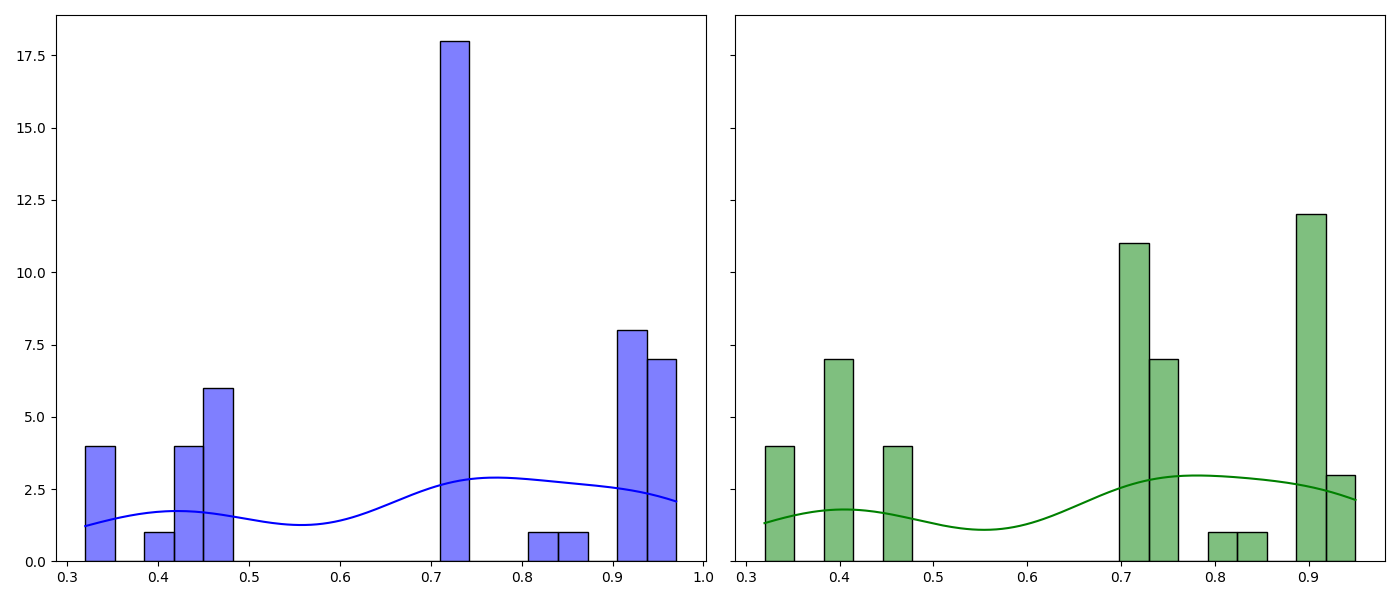}
    \caption{mux6}
  \end{subfigure}
  \caption{Comparison of RBO values. Blue (resp.~green) shows
    comparison with (resp.\ absolute) $\shap{E}$ scores} \label{fig:cmp}
\end{figure*}

SHAP and nuSHAP were compared using two metrics: i) the ranking of
feature importance imposed by different scores;
and ii) the runtime for computing these scores.
Specifically, for each tested instance, we first compute its nuSHAP
and SHAP scores.
We then determined the order of feature importance based on these
scores.
For SHAP, we considered two orders: one based on the original SHAP
scores and the other based on the absolute values of the SHAP scores. 
Next, we compare the order imposed by nuSHAP scores with the order
imposed by original SHAP scores, and separately compare the order
imposed by nuSHAP scores with the order imposed by the absolute values
of SHAP scores.

To compare the rankings of feature importance,
we used the metric rank-biased overlap
(RBO)~\cite{webber2010similarity} for each pair of scores.
RBO is a metric used to measure the similarity between two ranked
lists, and it ranges between 0 and 1.
A higher RBO value indicates a greater degree of similarity between
the two rankings, with 1 denoting a perfect match for the top-ranked
elements considered.
A publicly available
implementation\footnote{\url{https://github.com/changyaochen/rbo}.} of
RBO was used in our experiments.
Given the fact that most people focus on top-ranked
features~\cite{miller-pr56}, we set \emph{persistence} to $0.5$ and
\emph{depth} to $5$ in our setting. This means that we focused on the
top-5 features while placing greater emphasis on the top-3 features.
%
%
\cref{tab:rbo_sum} summarizes the RBO values for all the tested
instances, including the minimum, maximum, and mean RBO values.
%
%
As can be observed, there is essentially \emph{no} correlation between
the $\nshap{E}$ scores obtained with the nuSHAP tool, and the
$\shap{E}$ scores obtained with the SHAP tool.
\cref{fig:cmp} shows the distribution of RBO values for the different 
instances for the datasets corral and mux6, both of which are purely
binary.
(The blue (resp.\ green) histograms depict the RBO values comparing
nuSHAP and SHAP (resp.\ the absolute value of SHAP). 
Surprisingly, the RBO measure is never 1, and most often it is
not even close to 1.
For values of the RBO metric close to 0, this means that the important
features reported by the tool SHAP are not the important features
reported by nuSHAP. And as argued in this paper, nuSHAP is
\emph{guaranteed} not to mislead.
For larger values of the RBO metric (e.g.\ around $0.8{\sim}0.9$),
most of the important features may be the same, but either their
relative order of importance is computed incorrectly by the tool SHAP,
or one of the features is assigned undue importance.
Finally, \cref{tab:time_tabular} presents the average running times
for computing the different scores,
and the average number of samples used for computing the nuSHAP
scores, i.e.\ the same samples that the tool SHAP also uses. As can be
observed, nuSHAP compares favourably with the tool SHAP.

\section{Discussion} \label{sec:conc}

Recent work~\cite{msh-cacm24,hms-ijar24} revealed
critical flaws with the definition of SHAP scores \cite{lundberg-nips17},
i.e. the proposed use of Shapley values~\cite{shapley-ctg53} in XAI.
Concretely, this work 
constructed 
classifiers for which 
the theoretical SHAP scores
convey misleading information.
This implies that the tool SHAP~\cite{lundberg-nips17} approximates
scores that can be misleading.
%
As a consequence, the identified flaws question the conclusions from 
thousands of papers that build on the tool SHAP.

Motivated by these negative results, a number of more recent
works proposed alternatives to the use of SHAP
scores~\cite{izza-aaai24,ignatiev-sat24}. Unfortunately, these
alternatives are not based on Shapley values, and so they can
also produce unsatisfactory results~\cite{lhams-corr24}.

In contrast, this paper proposes a novel solution, one that targets
the efficient computation of a novel definition of SHAP scores,
thereby addressing the known flaws of the SHAP scores in current
use~\cite{msh-cacm24,hms-ijar24,lhms-aaai25}.
%
The experiments confirm the scalability of the proposed solution,
and
provide further evidence of SHAP's flaws, not only in theory, but also
in practice.
Concretely, the experiments demonstrate that the measures of feature
importance obtained with the tool SHAP are misleading in almost all
of the tests considered. The lack of quality in the rankings of the
scores obtained with the tool SHAP is always observable, and most
often very significant, confirming that these scores can be 
fairly unrelated with those obtained with the novel non-misleading 
definition of SHAP scores.

%
\begin{acks}
  This work was supported in part by the Spanish Government under
  grant PID2023-152814OB-I00, and by ICREA starting funds.
  This work was supported in part by the National Research Foundation, Prime Minister’s Office, 
  Singapore under its Campus for Research Excellence and Technological
  Enterprise (CREATE) programme.
  %
  %
  %
  This work was motivated in part from discussions with several
  colleagues including
  R.\ Passos,
  J.\ Planes, A.\ Morgado, R.\ Bejar,
  C.\ Menc\'{\i}a,
  L.~Bertossi, A.~Ignatiev, M.~Cooper and Y.~Izza.
  %
  %
\end{acks}


\newtoggle{mkbbl}

\settoggle{mkbbl}{false}

\bibliographystyle{ACM-Reference-Format}

\iftoggle{mkbbl}{
    \bibliography{refs}
}{

}

%

\end{document}